# A NOVEL ARCHITECTURE FOR 3D MODEL IN VIRTUAL COMMUNITIES FROM DETECTED FACE

Vibekananda Dutta
Central University of Rajasthan
Kishangarh, India

Dr.Nishtha Kesswani
Central University of Rajasthan
Kishangarh, India

Deepti Gahalot
Govt.Engineering
College, Ajmer

**ABSTRACT:**

Towards next generation, Criminal activity in virtual worlds is becoming a major problem for law enforcement agencies. Virtual communities such as Second Life will be quickly becoming the next frontier of cybercrime. Even now a day's Forensic investigators are becoming interested in being able to accurately and automatically track people in virtual communities. Mostly in the multimedia context, an avatar is the visual representation of the self in a virtual world. In this research paper we suggest how to extract a face from an image, modify it, characterize it in terms of high-level properties, and apply it to the creation of a personalized avatar. In this research work we tested, we implemented the algorithm on several hundred facial images, including many taken under uncontrolled acquisition conditions, and found to exhibit satisfactory performance for immediate practical use.

## 1. INTRODUCTION:

The term avatar, which refers to the temporary body a god inhabits while visiting earth. In virtual communities, it now describes the user's visual embodiment in cyberspace [1]. Virtual worlds are also extremely attractive for the run-of the-mill criminals interested in conducting identity theft, fraud, tax evasion, illegal gambling and other traditional crimes. In the virtual world increasingly populated by non-biological characters there are just no existing techniques for identity verification of intelligent entities other then self-identification.

Art metrics, which is defined as the science of recognition, detection and verification of intelligent software agents and industrial robots and other non-biological entities aims to address this problem. This future oriented sub-field of security has broad applications in this virtual world [2]. Artificially Intelligent programs are quickly becoming a part of our daily life. In this paper we suggest utilization of face detection systems and development of novel face recognition algorithms for face-based avatar creation.

## 2. FACE RECOGNITION

### 2.1 HUMAN FACE RECOGNITION PROCESS

Human faces have similarities and differences. They have a consistent structure and location of facial components (i.e. the relationship among eyes, nose, etc.).

In human face recognition where we have four stages [5]:

**a) Acquiring a sample:** In a complete, full implemented biometric system, a sensor takes an observation. The sensor might be a camera and the observation is a snapshot picture. In our system, a sensor will be ignored, and a **2D** or **3D** face picture "observation" will supplied manually.

**b) Extracting Features:** For this step, the relevant data is extracted from the predefined captured sample. This is can be done by the use of software where many algorithms are available. The outcome of this step is a biometric template which is a reduced set of data that represents the unique features of the enrolled user's face.

**c) Comparison Templates:** This depends on the application at hand. For identification purposes, this step will be a comparison between a given picture for the subject and all the biometric templates stored on a database. For verification, the biometric template of the claimed identity will be retrieved (either from a database or a storage medium presented by the subject) and this will be compared to a given picture.

**d) Declaring a Match:** The face recognition system will return a candidate match list of potential matches. In this case, the intervention of a human operator will be required in order to select the best fit from the candidate list. An illustrative analogy is that of a walk-through metal detector, where if a person causes the 7 detector to beep, a human operator steps in and checks the person manually or with a hand-held detector.

2.2 HUMAN FACE RECOGNITION TECHNIQUES

All available face recognition techniques can be classified into four categories based on the way they represent face [7];
**1.** Appearance based which uses holistic texture features.
**2.** Model based which employ shape and texture of the face, along with 3D depth information.
**3.** Template based face recognition.
**4.** Techniques using Neural Networks.

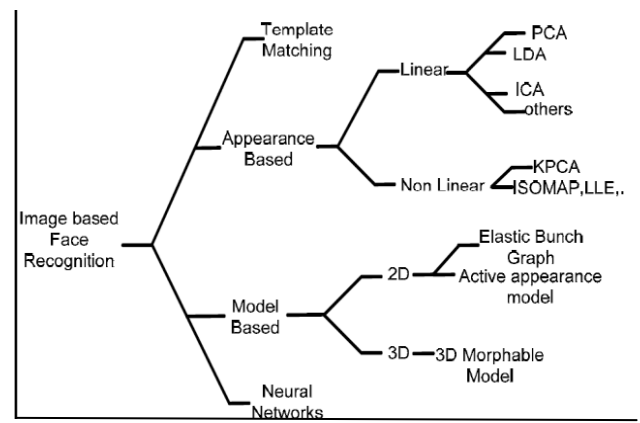

**Figure 1: Classification of Face Recognition Methods**

3. **AVATAR GENERATION FROM FACE RECOGNITION SYSTEM**

Avatar and human faces have similarities and differences. Both have a consistent structure and location of facial components (i.e. the relationship among eyes, nose, etc.). These similarities gives idea of an avatar face recognition framework designed in the same manner as human Face Recognition systems.

Avatar faces span a wider range of colors than human faces, and the colors provide strong discrimination between identities (see Fig. 2).

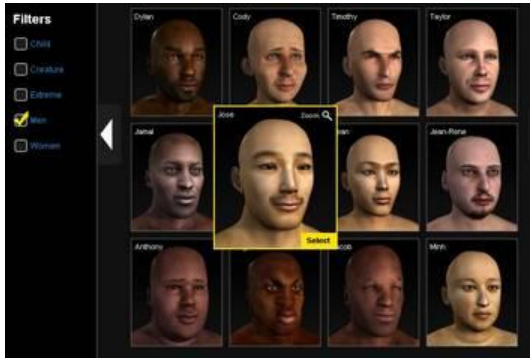

**Figure2: Examples of the different subjects in the Second Life avatar dataset. Each image corresponds to one of the different pose sets. In our matching experiments, the frontal image from group A was used as the gallery image. The remaining sets were all used as probe images**

We propose an algorithm for avatar generation from face recognition that follows the same procedures as standard face recognition systems [6], consisting of three stages:
1. Face detection and image normalization
2. Face representation
3. Matching and

In the last step which we proposed an Avatar generation system which produced 3D model of Avatar faces for that detected faces.

A. Face Detection and Image Normalization

In a virtual world, real-time face detection will detect the presence of an avatar subject with a frontal to near-frontal face in the field of view [9], similar to traditional face recognition.

Once an avatar face is detected it must be pre-processed by performing both geometric and color normalization in order to reduce.

Variations caused by external parameters such as camera location and illumination. We found a method to work with similar effectiveness on avatar faces. We explored the use of (i) a Morph able feature based extraction for trained on avatar faces, and (ii) the default Morphable feature based extraction packaged with OpenGL & MatLab [8].

B. Face Representation

In order to match two faces in Avatar face recognition, we represent the face in a metric space by first computing a set of local feature descriptors across the face region. Two separate feature descriptors are used to describe (i) the structure of the face, and (ii) the appearance properties of the face. For computing the local descriptors, the normalized face image is divided into an ordered set of N overlapping square patches $P_i$, i = 1.... N, each of size $S_p \times S_p$, $S_p$ = 32. For each patch $P_i$ two feature vectors are extracted: one describing the appearance $(A_i \in R^{da})$, and the other describing the structure $(A_i \in R^{ds})$ [9, 10], Computing features across a set of overlapping patches allows for salient descriptions at specific locations of the face that is robust to variations in geometric normalization.

C. Matching

For a given avatar face, we have two sets of vectors $S_i$ and $A_i$, i = 1..... N, where N is the number of face is patches. To determine an avatar's identity, we first concatenate the set of local (patch) descriptors into a single feature vector of length **Nd$_s$** and **Nd$_a$**, respectively for **S$_i$** and **S$_a$**. The concatenated feature vectors are represented as **S$^j$** and **A$^j$** for the **j**-th avatar subject. The distance between two faces corresponding to images **m$_p$** and **m$_j$** computed using cosine correlation similarity measure given by [11]:

$$\cos(m_p, m_j) = \frac{m_p^t m_j}{|m_p||m_j|} \quad (1)$$

Similarity between two labelled graphs is the average of this vector similarity over corresponding nodes of the facial graph. This is essentially a measure of the filter response amplitude at corresponding spatial frequency, orientation, and position on the grid.

### D. An Avatar Generation System

An avatar creation system which we consider to be desirable is the ability to embody some intelligence about the face being processed. An automatic semantic analysis of the face into facial categories empowers the system to generate intelligent suggestions of avatar body type [3].

An important design feature of the Avatar Creation system is rapidity of uses. In a few seconds a user's picture is filtered, the region containing the face is extracted from the image and registered with an adaptable grid, facial attributes are classified and semantic labels attached to the face and finally the system suggests an interesting looking avatar body to the user.

## 4. EXPERIMENT AND DISCUSSION

### a) Data

For the purpose of 3D model of facial generation, various type of virtual world and 3D model creation softwer were consider based on the use of this research work, including[12].

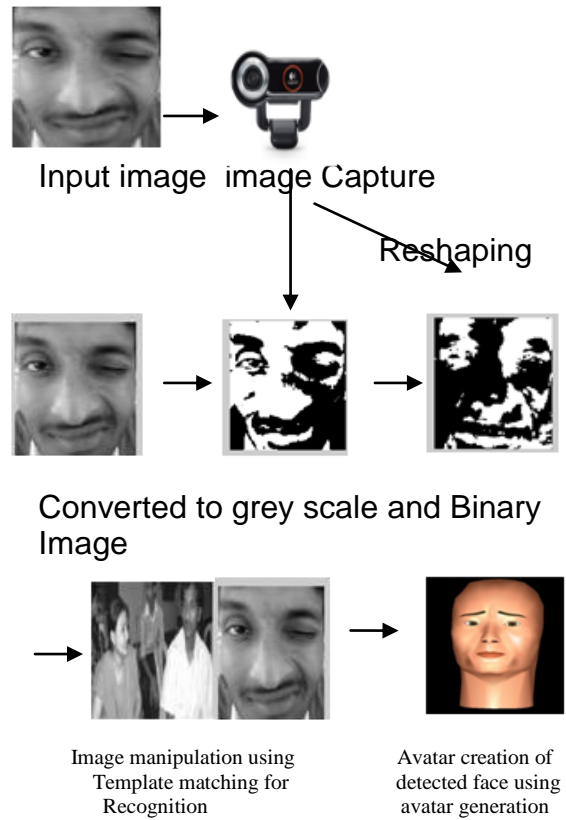

Input image   image Capture

Reshaping

Converted to grey scale and Binary Image

Image manipulation using Template matching for Recognition    Avatar creation of detected face using avatar generation

**Figure 3: Functional modules of the Avatar generation system**

1. ability to view the 3D model facial from different angles.
2. Selecting the constructing facial featuree in generation new Avatar faces.
3. Mutable attributes to 3D facial features.

**b)** 3D model avatar Maker

Before having implemented the 3D model Avatar generation with the original facial dataset, we 1st implemented our experiment on the images from the university given datasets [13], where we converted the **given** images to 3D-Avatar using the **openGL** programming according to our experiment, where the 3D model Avatar maker shows the user to make his own 3D model of Avatar for second life from a simple images.

c) Result

The images of **100** persons with different angles makes the number of images near about **800-900** images, where we grouped those images in to **3** datasets each of which near about **300** images using Morphable model based template matching technique for face recognition. Where we first detect the images then from that detected images we generate the 3D facial character.

**Table: Performance comparison of different dataset with avatar generation system**

| Set | Number of Images | 3D Morphable Template Matching Technique | Avatar Generation from detected faces using openGL programming for 3D-avatar creation |
|---|---|---|---|
| A | 300 | 80.8% | 84% |
| B | 270 | 89.2% | 91% |
| C | 281 | 72.9% | 81% |

Using Morphable model based face recognition [14], a Rank-1 accuracy of 97.58% was achieved. Testing the prototype avatar system proposed on this dataset was not warranted because: (i) A Morphable based face matcher already achieves a very high accuracy, and (ii) the proposed system is designed to match images to avatars using less realistic renderings, such as those in Second Life.

These results are reported because it is useful to know that current face recognition technology appears to be sufficiently accurate in the 3D model architecture, when the 3D-model is rendered using advanced software programming such as openGL.

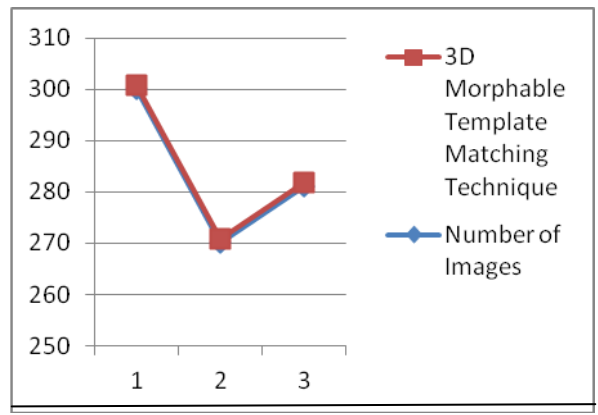

**Figure 4: Graphical representation of comparison in different dataset**

## 5 CONCLUSIONS

This Research paper addressed the problem of generating the 3D-model faces from face recognition system. We have reported results of experiments aimed towards the Potential directions for future research include the investigation of other visual and behavioural approaches to virtual world security based on appearance of new characteristics and abilities in the 3D-model. As virtual reality technology progresses day by day and criminal activity become the major problem, so it will require new security solutions for identity management across worlds populated by both human and artificial entities [13].

## 6. ACKNOWLWDGEMENT

**REFERENCES**

[1] B. Damer. Avatars! Exploring and Building Virtual Worlds on the Internet. Peachpit Press,
Berkeley, 1997.

[2] O'Harrow, R., *Spies' Battleground Turns Virtual*, in The Washington Post. February 6, 2008: Available at: http://www.washingtonpost.com/wpdyn/content/article/2008/02/05/AR2008020503144.html.

[3] Yampolskiy, R. V. and V. Govindaraju (2008). Behavioral Biometrics for Verification and Recognition of Malicious Software Agents. Sensors, and Command, Control, Communications, and Intelligence (C3I) Technologies for Homeland Security and Homeland Defense VII. Orlando, Florida.

[4] S.Inoue, M. Ishiwaka, S. Tanaka. & J. Park. An image Expression Room. IEEE Proceedings of the International Conference on Virtual Systems and Multimedia VSMM 97 p.181, 1997.

[5] Statistics in Face Recognition: Analyzing Probability Distributions of PCA, ICA and LDA Performance Results Kresimir Delac 1, Mislav Grgic 2 and Sonja Grgic 2 *1 Croatian Telecom, Savska 32, Zagreb, Croatia, e-mail:* kdelac@ieee.org *2 University of Zagreb, FER, Unska 3/XII, Zagreb, Croatia*

[6] Lyons, M., et al., Avatar Creation using Automatic Face Recognition, in ACM Multimedia 98. Sept. 1998: Bristol, England. p. 427-434.

[7] Evaluation of Face Recognition Techniques for Application to Facebook, Brian C. Becker Carnegie Mellon Univ 5000 Forbes Av Pittsburgh, PA 152 briancbecker@cmu.

[8] Open source graphical library and math works library.

[9] Viola, P. and M.J. Jones, Robust real-time face detection. Int. Journal of Computer Vision, 2004. **57**: p. 137-154.

[10] Ahonen, T., A. Hadid, and M. Pietikainen, Face description with local binary patterns: Application to face recognition. IEEE Trans. Pattern Analysis and Machine Intelligence, 2006. **28**: p. 2037- 2041

[11] Ojala, T., M. Pietikainen, and T. Maenpaa, *Multiresolution grayscale and rotation invariant texture classification with local binary patterns.* IEEE Trans. Pattern Analysis & Machine Intelligence, 2002. **24**: p. 971-987.

[12] Oursler, J.N., M. Price, and R.V. Yampolskiy, Parameterized Generation of Avatar Face Dataset, in 14th InternationalConference on Computer Games: AI, Animation, Mobile, Interactive Multimedia, Educational & Serious Games. 2009: Louisville, KY.

[13] Phillips, P.J., et al., The FERET evaluation methodology for face recognition algorithms. IEEE Transactions on Pattern Analysis and Machine Intelligence, October 2000. **22(10)**: p. 1090-1104.